%% file: author.tex
\documentclass{styles/svproc}
\usepackage{url}
\usepackage{graphicx}
\usepackage{subcaption}

\usepackage{multirow}

\begin{document}
\mainmatter              
\title{Hostility Detection and Covid-19 Fake News Detection in Social Media}
\titlerunning{Hostility Detection and Covid-19 Fake News Detection in Social Media}  
%
\author{Ayush Gupta\inst{*} \and Rohan Sukumaran\inst{1}\inst{*} \and
Kevin John\inst{2}\inst{*} \and Sundeep Teki }
\authorrunning{Gupta, Sukumaran, John and Teki} 
%
\tocauthor{Ayush Gupta, Rohan Sukumaran, Kevin John, Sundeep Teki}
%
\institute{PathCheck Foundation, Cambridge 
\and
Indian Institute of Information Technology, Sri City, India 
\\
\email{ayush.goc@gmail.com}, \email{\{rohan.s16, kevin.j17\}@iiits.in},\\ \email{sundeep.teki@gmail.com},
}

\maketitle              
\footnotetext[1]{*Denotes equal contribution.}
\begin{abstract}
With the advent of social media, there has been an extremely rapid increase in the content shared online. Consequently, the propagation of fake news and hostile messages on social media platforms has also skyrocketed. In this paper, we address the problem of detecting hostile and fake content in the \textit{Devanagari} (Hindi) script as a multi-class, multi-label problem. Using NLP techniques, we build a model that makes use of an abusive language detector coupled with features extracted via Hindi BERT and Hindi FastText models and metadata. Our model achieves a 0.97 F1 score on coarse grain evaluation on Hostility detection task. Additionally, we built models to identify fake news related to Covid-19 in English tweets. We leverage entity information extracted from the tweets along with textual representations learned from word embeddings and achieve a 0.93 F1 score on the English fake news detection task. 

\end{abstract}
\input{Introduction}
\input{relatedwork}

\input{Methodology}
\input{Experiments}
\input{Results}
\input{Conclusion}

\bibliographystyle{plain}
\bibliography{styles/bibtex/references}

\end{document}

%% file: Introduction.tex
\section{Introduction}
\label{Introduction}

With the proliferation of social media platforms, information dissemination knows no bounds. The public discourse that occurs on social media platforms leads to a lot of information being shared and consumed, that more often than not, cannot be verified as hostile free content or fact-checked for authenticity. Usually, such harmful content is used to defame and hurt people, especially from minority groups, specific countries, the LGBTQ community, and several other organizations.
Recently, as the COVID-19 pandemic is devastating humanity, hostile content and misinformation on social media have transformed into an “Infodemic”, further expediting the need for control of such content. This being said, fake news and hostility detection are daunting tasks owing to the context, domain knowledge, and the inherent complexity of the natural language constructs, like different forms of hatred, target groups, ambiguous word representations, and more, are required to understand it.


In this paper, as part of the Shared Task organized in CONSTRAINT 2020 \cite{bhardwaj2020hostility,patwa2021overview,patwa2020fighting}, we look at fake news detection for English and Hindi (\textit{Devanagari}) tweets, as well as hostility detection for Hindi tweets. We evaluated machine learning and deep learning models and propose an ensemble model that combines language representations with the features (linguistic and otherwise) extracted from the data. Furthermore, we measure the differential impact of each feature and also notice that our model was able to provide the correct annotation for a few of the miss-annotated samples in the English fake news dataset, opening up to possibilities of weak supervision for labeling. 

Our contributions in this paper are primarily the identification of entities pertinent to fake news detection and data analysis steps that can be taken to understand such micro-blogging data. We show that a lightweight ensemble of Word2Vec \cite{mikolov2013efficient} coupled with entity information gives better performance in English fake news detection than larger BERT \cite{devlin2018bert} based models. Our ensemble model achieved \textbf{0.93 F1 score} (above the baseline scores). Furthermore, we perform Hostility detection by ensembling models based on Hindi BERT and Hindi FastText. Further, we create various simple but effective text features extracted from the tweets to be fed in the model and also compile a list of abusive words that are used in an abusive words detector, which further boosts the accuracy. Our methods obtained a weighted F1 score of \textbf{0.97} in coarse-grain classification for Hostility task and ranked \textbf{5th} out of 45 teams. For the fine-grain classification, our model obtained a weighted F1 of \textbf{0.62} and ranked \textbf{7th} out of 45 teams.

%% file: relatedwork.tex
\section{Related Works}\label{relatedwork}


Riedel et al., (2017) \cite{riedel2017simple} used TF-IDF similarity scores between the title of the news and the content concatenated with the corresponding TF-IDF vectors. This was further fed into a fully connected neural network. Albeit impressive results, this approach is not useful for identifying fake news that is spreading via micro-blogging sites. Further, work from Khatter et al. (2019) \cite{khattar2019mvae} looks at solving the problem using a multimodal approach. The authors make use of a Variational Autoencoder to learn the representation of a social media post, combining the textual and image representations.
More recent work from Lee et al. (2020) \cite{lee2020language} utilizes a language model as a zero-shot fact-checker. A masked language model (MLM) training followed by entailment check is adapted. The model is tested in a closed-book setting. The results look promising but require the presence of large scale fact-based data to enable such training. Vijjali et al. (2020) \cite{vijjali2020two} looked at the problem of claim verification by following a two-stage pipeline. The paper uses a combination of BERT \cite{devlin2018bert} and ALBERT \cite{lan2019albert} models - for identifying explanations based on the claims from a large dataset and checking if the claims are entailment of the explanations.

Bohra et al. (2018)\cite{bohra2018dataset} uses various linguistic features such as character, word, and lexicon-based features on Hindi-English code-mixed tweets into Hate speech or Normal speech and attain 71.7\% accuracy with an SVM model. In \cite{xu2012learning} the authors use sentiment analysis to detect bullying in tweets and Latent Dirichlet Allocation (LDA) topic models \cite{blei2003latent}.
Also, hate speech detection is a widely studied topic \cite{djuric2015hate,kwok2013locate} with most of the work focused on developing binary classification models for English.

Research in Hostility detection is more focused on learning the discriminatory features via Deep Learning models instead of relying on hand-crafted features. Bhardwaj et al. (2020) \cite{bhardwaj2020hostility} computes input embedding with help of a pre-trained multilingual BERT(m-BERT) model\cite{devlin2018bert},and is then fed into several machine learning models like Random Forest (RF), SVM, and Linear Regression (LR). Hostility detection on \textit{Devanagari} is scant, and we apply Hindi FastText and mBERT embeddings for hostility detection in Hindi; the classes for fine-grain classification:  Hate, offensive, Defamation, Fake, and non-hostile, and coarse-grain classification: Hostile or non-hostile.

%% file: Methodology.tex
\section{Method}\label{Methodology}
\subsection{English Fake News}

\subsubsection{Preprocessing}
As the data in Patwa et al. (2020) \cite{patwa2020fighting} are based on tweets scraped from the internet - abbreviations, social media lingo, URLs amongst other tokens are present in the dataset. The dataset contained an almost 50-50 split of real and fake tweets in train, validation and test. In order to create embeddings from the data using Word2Vec\cite{mikolov2013efficient}, FastText\cite{bojanowski2017enriching}, BERT\cite{devlin2018bert} and many more, we used multiple preprocessing steps. The data was initially rid of contractions, for instance  - y’all → you all, how’re → how are,and so on.
\begin{figure}[]
    \includegraphics[width=12cm]{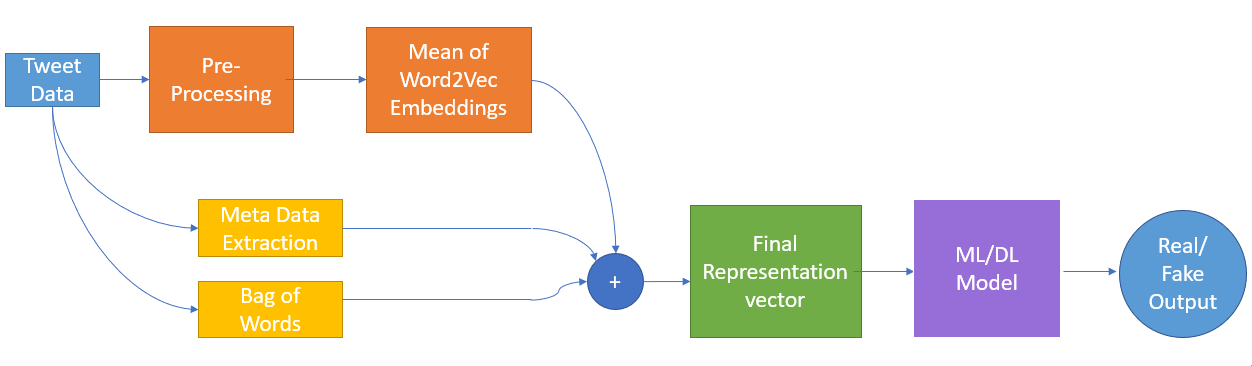}
    \caption{Data and modeling pipeline for English Fake News detection}
    \label{fig:pipeline-fake}
\end{figure}
These contractions were identified through a combination of regex pattern matches and simple rules. Further, occurrences of URLs, mentions, emojis, numbers and so on were cleaned and removed with the help of tweet-preprocessor\footnote{https://pypi.org/project/tweet-preprocessor/} library, the punctuation and special characters were stripped and the remaining text was tokenized with the help of NLTK libraries.

\subsubsection{Rule-based models}
Based on data analysis we observed that length of tweets, number of hashtags, number of mentions, etc. had a statistically significant impact in the class labels.  All such similar patterns specific to the dataset were analysed and added as features to the model.

\subsubsection{Distributional embeddings}
Using distributional word embeddings like Word2Vec \cite{mikolov2013efficient}, FastText \cite{bojanowski2017enriching} etc. has seen huge success in text classification. Word2Vec being a word level embedding suffers from the issue of Out Of Vocabulary (OOV) distributions. To mitigate this, an often adapted strategy is making use of sub-word level embeddings like FastText, BERT and more. All these models were used to convert raw text into a fixed-length embedding vectors.

\subsubsection{Adding entity-based information using Bag of Words (BoW)}
The data has significant presence of entities like names of government bodies, scientific journals, places, organizations, etc and the distribution varies per class. 
Using TF-IDF based rankings, we built a Bag of Words (BoW) feature vector. This would indicate the presence or absence of each of the corresponding terms in a given sentence.

\subsubsection{Contextual embeddings - BERT based models}
Transformer models \cite{vaswani2017attention} like BERT \cite{devlin2018bert}, RoBERTa \cite{liu2019roberta}, ALBERT \cite{lan2019albert}, ELECTRA \cite{clark2020electra}, have shown success on NLP tasks like text classification, question answering sequence labeling and more. We make use of these models to find meaningful representations of the input sentences. 

\subsubsection{Ensemble model}
 The idea behind the ensemble technique is to leverage the performance and robustness captured by the different models. Ensembling can be done using a range of methods from using a logical OR/logical AND operation over the “n” models’ predictions, to using fully connected neural network layers to learn the representations of the models. Here, we also combine the metadata (length of tweet, number of mentions, etc.) and Bag of Words along with the predictions from different models, into a single vector representation and fed into a fully connected neural network for ensembling. The overview of the data pipeline is mentioned in Figure \ref{fig:pipeline-fake}.

\subsection{Hostility and Fake News Detection Hindi}

\subsubsection{Problem Formulation} 
 The Devanagari dataset demanded a coarse grained evaluation (Hostile or not) as well as a fine grained one(Non-Hostile, Hate, offensive, Fake and Defamation).
\subsubsection{Data and Preprocessing}

Data distribution can be seen from Table \ref{tab:data-hindi}. Prior to training the models, we perform a few preprocessing steps as follows:
\begin{table}[!t]
\centering
\begin{tabular}{|c|c|c|c|c|c|c|}

\hline
           & \multicolumn{5}{c|}{\textbf{Hostile posts}}                                           & \textbf{Non-Hostile} \\ \hline
           & \textbf{Fake} & \textbf{Hate} & \textbf{Offensive} & \textbf{Defame} & \textbf{Total} & \textbf{}            \\ \hline
Train      & 1144          & 792           & 742                & 564             & 2678           & 3050                 \\ \hline
validation & 160           & 103           & 110                & 77              & 376            & 435                  \\ \hline
Test       & 334           & 237           & 219                & 169             & 780            & 873                  \\ \hline
Overall    & 1638          & 1132          & 1071               & 810             & 3834           & 4358                 \\ \hline
\end{tabular}
\caption{The data distribution for Hindi Hostility Dataset.}
\label{tab:data-hindi}
\end{table}

\begin{itemize}
    \item \textbf{Tokenization and removal}:  We designed a tweet-tokenizer to parse and remove every username-mention, hashtags, and URLs present.
    \item \textbf{Stopword Removal}: Tokenized tweets were passed through a customized stop word removal for the Hindi language.
    \item \textbf{Emoji Removal}: Demoji library\footnote{https://pypi.org/project/demoji/} was used to identify and remove emojis.
    \item \textbf{Tweet cleaning}:  Removes all punctuation, new line, Unicode characters (u200B, u200C, u200D) which are zero-length white spaces.
\end{itemize}
 
\subsubsection{Baseline Classifier}
As baselines, we experiment with three broad representations.
(1)TF-IDF (2)mBERT Embedding  (3)FastText Embedding 
\subsubsection{TF-IDF feature}
TF-IDF as the basic approach for text representation is widely used to score words thus helping in applications like text classification, topic modeling, etc. We clean the tweets by passing them through the above four preprocessing steps. 
The feature vector created from the clean tweets is passed into the machine learning models.

\subsubsection{mBERT Embedding}
The initial baseline was defined by using the multilingual BERT (m-bert) model \cite{libovicky2019language} to handle the Hindi data. We use the approach defined by the authors\cite{bhardwaj2020hostility} to replicate the results for baseline. 
The preprocessing was a four-step process, defined in the previous subsection.
We calculate the word embedding for each sentence in a tweet by tokenizing it and passing through the embedding matrix defined using the mBERT model. We use sentence embeddings as a feature for machine learning models.

\subsubsection{FastText Embedding}
FastText\footnote{https://fasttext.cc/} is an open-source, lightweight library trained on 157 different languages, that allows users to learn text representations and text classifiers. FastText treats each word as a composition of character n-grams thereby the word representation is created by the sum of the character n-gram representations.
The method of calculating sentence embedding is the same as described in the previous section. 



\subsubsection{Feature Identification and Extraction}
The feature Extractor module is used to identify and extract linguistic features from a given natural unstructured tweet. The feature Extraction module leverages several natural language processing (NLP) and regex techniques to extract these features. We have categorized them broadly into the following 3 modules: a) Abusive Language Detector b) Username Mention, URL, and Hashtag count c) Emoji Detector. 

\subsubsection{Abusive Language Detector}


We hypothesize that giving a count of abusive words as a feature to the classification model boosts the accuracy in detecting offensive tweets. To validate our hypothesis, we manually curated a list of commonly used 84 profane words in Hindi. We measure the presence of these profane words per tweet and use this count as a feature for the classification model. 

\subsubsection{Username Mention, URL, and Hashtag counter}
On analyzing the dataset, we found that on an average there were more hashtags and URLs  in the non-hostile tweets as compared to the hostile ones, whereas there are more user mentions in the Hate tweets. Informed by this, we hypothesize that the user mention, presence of URL or Hashtag can have differential impact  on the fine-grain classes (Hate, Offensive, Fake, and Defamation).
Therefore we design a rule-based method to parse each tweet to extract the count of the occurrence of these entities and use the count as an input feature.

\subsubsection{Emoji Detector}
The inherent nature of micro-blogging(short messages) has made people use acronyms, emoticons, and various other special characters to express their message. 
 We hypothesize that the use of emoji as a feature for hostility detection could increase the accuracy, as similar results were observed in sentiment analysis\cite{wolny2016emotion}. To validate this hypothesis, we use the demoji library to extract all the emojis present in the tweet and pass the count corresponding to each tweet as a feature to the classification model along with other features.
\begin{figure}[]
    \centering
    \includegraphics[width=12cm]{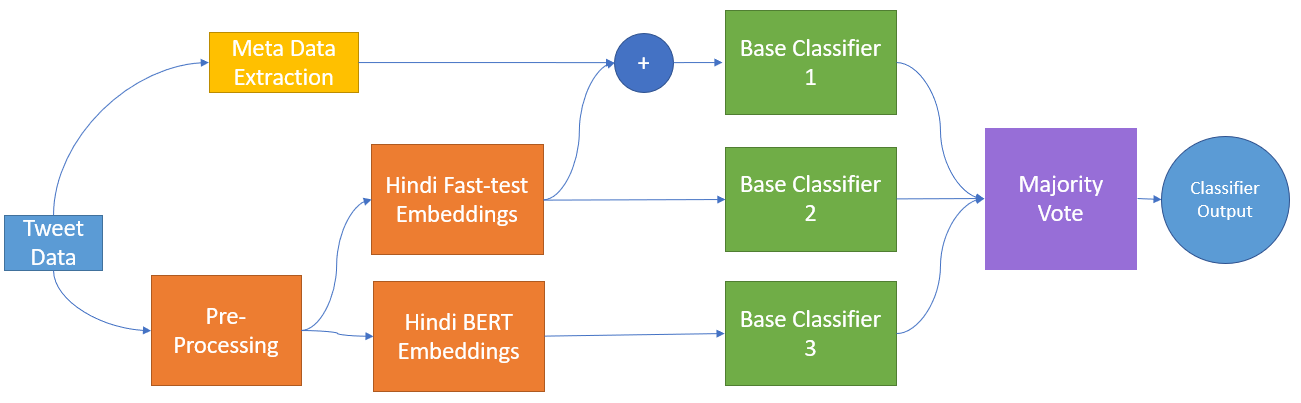}
    \caption{Data and modeling pipeline for Hindi Hostility Classification}
    \label{fig:pipeline-hostility}
\end{figure}

\subsubsection{FastText based LSTM + FastText Classifier}

FastText represents a text by average of word vector and it also allows the word vector to be updated through back-propagation while training. 
FastText classifier uses (multinomial) logistic regression for training on the vector average and leverages Hierarchical Softmax to reduce computational complexity. We pass the cleaned training data to the FastText, and use grid search to find the optimal hyperparameter for the model. 
For the multi-label classification we use one-vs-all strategy.  
\subsubsection{FastText based LSTM + Features}

In this paper, we built the hostility classiﬁcation model using the LSTM\cite{hochreiter1997long} model with FastText embedding and metadata. We use the LSTM model owing to its internal memory to process long-range dependencies in sequential inputs like tweets. This model takes two inputs, 
the first being V = \{v1, v2, ..., vn\} the FastText Embedding, where each v\textsubscript{i} represents a cleaned tweet  in vector space and the second  MD =  \{A, E, H, M, U\} the 5-tuplet metadata representation where A, E, H, M, U are the count of abusive words, emoji,  hashtag, mention, and URL per clean tweet respectively. 
We use keras\footnote{https://keras.io/} to define and train our model.
\subsubsection{BERT-Hindi Embedding with BERT model} 
Recent papers using BERT based contextual embedding models have shown SOTA performance in Hostile language identification\cite{mozafari2019bert,nikolov2019nikolov}.
We use Hindi-BERT pre-trained model\footnote{https://huggingface.co/models} and fine-tune it on our dataset. The Hindi BERT model has 12 hidden layers, with an encoding layer of 256 dimensions instead of 786, the model has 4 attention heads for each attention layer. The dimension of the feed-forward layer is kept as 1024 instead of 3072 which is the default in BERT models.
We find the best learning rate (i.e., 1.2e-4 here) for the model by simulating the training data briefly with different learning rates and choose the one with minimum loss. For training, we pass the preprocessed tweets into the BERT model and train the model for 12 epochs with a batch size of 8. 

\subsubsection{Ensemble Model}
In this approach, we ensemble three base classifier, which are FastText Embedding with FastText classifier, FastText Embedding with LSTM model, and Hindi BERT Embedding with BERT model. The whole framework of ensemble setting is illustrated in Fig. 2.  Each of the base classifier is explained in the above subsection. The final classification is made by fusing the outputs of all three base classifiers through majority voting. 

%% file: Experiments.tex
\section{Experiments}\label{Experiments}

\subsection{English Fake News}
\subsubsection{Models}
We experimented by using the word embeddings extracted from FastText and Word2Vec and we used the mean of these word embeddings of the tweet as input to the machine learning models, as shown in Figure \ref{fig:pipeline-fake}. By examining the predictions of our best models, we noticed a pattern of misclassification. In order to mitigate this, we include entity level information extracted from the data.
\subsubsection{TF-IDF scoring for entities}
We used a scoring factor to determine useful words to be included in the BoW method by first calculating the tf-idf of misclassified tweets in the validation data separated by the class(real/fake), and then comparing those scores with the TF-IDF scores in the train data separated by class(real/fake). For a real Tweet misclassified in validation data, let V\textsubscript r be the TF-IDF score of a word among the real misclassified tweets. Let T\textsubscript r and T\textsubscript f be the TF-IDF score of a word from the real and fake tweets in train respectively. Further, we find words which have high scores for V\textsubscript r and T\textsubscript r while simultaneously have low scores of T\textsubscript f to included these words to the BoW feature vector. 
\subsubsection{Contextual Embedding}
Owing to the success of BERT based models in a suite of NLP tasks, we calculated the contextual embeddings constructed from BERT. This embeddings, concatenated with the meta features as shown in Fig \ref{fig:pipeline-fake}, was further fed as input to the machine learning or deep learning models. 

\subsection{Hindi Hostility Detection and Fake News}
We set up our experiment in three parts. Firstly, for baseline accuracy, embeddings - \textit{TF-IDF, mBERT, and Hindi FastText} - and models - \textit{RF, LR, SVM, Gradient Boosting(GB), MLP} - are set up. We use grid search for hyperparameter tuning. The results of the baseline classifiers are shown in Table \ref{tab:hate-results} in the TF-IDF, mBERT, FastText sections.

Secondly, we do an ablation study to determine the effectiveness of the linguistic features extracted from tweets. For this, we chose FastText Embedding as the base classifier and conducted a combination of experiments utilizing m1, m2, m3 metadata. The results are shown in Table \ref{tab:hate-results}.

Finally, we test our proposed models that are FastText Classifier(FC),  BERT-Hindi Embedding with the BERT model, FastText Embedding with LSTM plus all metadata. Further, conducted experiments with the combination of the an ensemble model, due to limited space we only show the best ensemble model in the Table \ref{tab:hate-results} in the Proposed Model section
All the experiments for coarse-grain and fine-grain are conducted in two parts, one on the validation dataset, and the other on the testing dataset and weighted F1 score is used as evaluation metrics.

%% file: Results.tex
\section{Results and Discussions}\label{Results}

\subsection{English Fake News Result}
As seen from Table \ref{tab:table-fake-eng}, the model which makes use of the metadata and BoW features along with embeddings and pseudo labeling gives the best performance. We hypothesize and validate the presence of entities involved in hostility and fake news detection. Adding such entities via a BoW feature vector increases the performance of the model. We also notice that contextual embeddings based on BERT architecture didn't provide huge improvements and in fact performed worse than our best model. This could possibly due to the nature of the dataset and the annotation strategy adapted. Furthermore, it is also important to note that there were a few samples in the English fake news dataset which were mis-annotated, but were correctly labeled by our model.

\begin{table}[!t]
\centering
\begin{tabular}{|l|l|l|}

\hline
\textbf{Model }                            & \textbf{Val Score (\%)}  & 
\textbf{Test Score (\%)}  \\
\hline
\textbf{Meta with SVM}                                             & 93.59                            & \textbf{93.45}   \\ \hline
\textbf{Meta + BoW with SVM}               & \textbf{93.74}   & 93.41 \\ \hline
Meta + BoW Ensemble OR                                    & 91.86                            & 91.66  \\ \hline
Meta + BoW Ensemble Majority Vote & 93.55    & 93.10  \\ \hline
Meta + BoW with SVM Pseudo-labelling                      & 93.64 & 93.41 \\\hline

\end{tabular}
\vspace{1 pt}
\caption{The weighted F1 scores on the English Fake news dataset. We notice that simple distributed embedding based models coupled with meta-data and entity information (via BoW features) exhibit the best performance. The \textbf{pseudo-labelling} was done on the test data}
\label{tab:table-fake-eng}
\end{table}


\begin{table}[!htb]
\begin{tabular}{|c|c|c|c|c|c|}

\hline
\multirow{2}{*}{\textbf{Method}}                                                           & \multirow{2}{*}{\textbf{Models}}         & \multicolumn{2}{c|}{\textbf{\begin{tabular}[c]{@{}c@{}}Coarse Grain \\ w.F-1\end{tabular}}} & \multicolumn{2}{c|}{\textbf{\begin{tabular}[c]{@{}c@{}}Fine Grain\\ w.F-1\end{tabular}}} \\ \cline{3-6} 
                                                                                           &                                          & Validation                                   & Test                                         & Validation                                  & Test                                       \\ \hline
\multirow{5}{*}{\begin{tabular}[c]{@{}c@{}}TF-IDF \end{tabular}} & TF-IDF + RF                              & 0.78                                         & 0.77                                         & 0.35                                        & 0.34                                       \\ \cline{2-6} 
                                                                                           & TF-IDF + SVM                             & 0.79                                         & 0.80                                         & 0.34                                        & 0.36                                       \\ \cline{2-6} 
                                                                                           & TF-IDF + LR                              & 0.76                                         & 0.75                                         & 0.33                                        & 0.32                                       \\ \cline{2-6} 
                                                                                           & TF-IDF + GB                              & 0.79                                         & 0.79                                         & 0.34                                        & 0.35                                       \\ \cline{2-6} 
                                                                                           & TF-IDF + MLP                             & 0.71                                         & 0.69                                         & 0.29                                        & 0.29                                       \\ \hline
\multirow{5}{*}{\begin{tabular}[c]{@{}c@{}}mBERT based\end{tabular}}  & mE +RF                      & 0.79                                         & 0.80                                         & 0.53                                        & 0.51                                       \\ \cline{2-6} 
                                                                                           & mE + SVM                    & 0.84                                         & 0.84                                         & 0.54                                        & 0.54                                       \\ \cline{2-6} 
                                                                                           & mE + LR                     & 0.83                                         & 0.84                                         & 0.51                                        & 0.51                                       \\ \cline{2-6} 
                                                                                           & mE + GB                     & 0.81                                         & 0.81                                         & 0.52                                        & 0.52                                       \\ \cline{2-6} 
                                                                                           & mE + LSTM                   & 0.83                                         & 0.83                                         & 0.54                                        & 0.53                                       \\ \hline
\multirow{5}{*}{\begin{tabular}[c]{@{}c@{}}FastText based\end{tabular}}  & FE+ RF                   & 0.83                                         & 0.82                                         & 0.51                                        & 0.50                                       \\ \cline{2-6} 
                                                                                           & FE+ SVM                  & 0.87                                         & 0.87                                         & 0.55                                        & 0.55                                       \\ \cline{2-6} 
                                                                                           & FE+ LR                   & 0.85                                         & 0.86                                         & 0.52                                        & 0.52                                       \\ \cline{2-6} 
                                                                                           & FE+ GB                   & 0.85                                         & 0.84                                         & 0.52                                        & 0.53                                         \\ \cline{2-6} 
                                                                                           & FE+ LSTM                 & 0.91                                         & 0.91                                         & 0.56                                        & 0.56                                       \\ \hline
\multirow{5}{*}{\begin{tabular}[c]{@{}c@{}}Meta Data Ablation\end{tabular}}              & FE + LSTM + m1           & 0.92                                         & 0.92                                         & 0.58                                        & 0.58                                       \\ \cline{2-6} 
                                                                                           & FE + LSTM + m2           & 0.92                                         & 0.92                                         & 0.57                                        & 0.58                                       \\ \cline{2-6} 
                                                                                           & FE + LSTM + m3           & 0.93                                         & 0.93                                         & 0.58                                        & 0.58                                       \\ \cline{2-6} 
                                                                                           & FE + LSTM + m1 +m2       & 0.92                                         & 0.92                                         & 0.58                                        & 0.58                                       \\ \cline{2-6} 
                                                                                           & FE + LSTM + m1 +m3       & 0.94                                         & 0.93                                         & 0.59                                        & 0.60                                       \\ \cline{2-6} 
                                                                                           & FE + LSTM + m2 +m3       & 0.93                                         & 0.93                                         & 0.58                                        & 0.58                                       \\ \hline
\multirow{5}{*}{Proposed Model} & FE+ LSTM + m1 + m2 + m3  & 0.94  & 0.95 & 0.59 & \textbf{0.63}\\ \cline{2-6} 
           & FE + FC & 0.92                                         & 0.92                                         & 0.58                                        & 0.58                                       \\ \cline{2-6} 
                                                                                           & HBE + Bert        & 0.96                                         & 0.96                                         & 0.60                                        & 0.61                                       \\ \cline{2-6} 
                                                                                           & Ensemble Model                           & \textbf{0.97}                                & \textbf{0.97}                                & \textbf{0.61}                               & 0.62                                       \\ \hline
\end{tabular}
\vspace{0.15 pt}
\caption{The results on the Hindi Hostility dataset. Here m1, m2, m3 are abusive language count, URL, mention, Hashtag count, and emoji count, respectively. FE, mE, HBE, are FastText, mBERT, and Hindi-BERT Embedding. }
\label{tab:hate-results}
\end{table}

\begin{figure}
\centering
\begin{subfigure}{.5\textwidth}
  \centering
  \includegraphics[width=.8\linewidth]{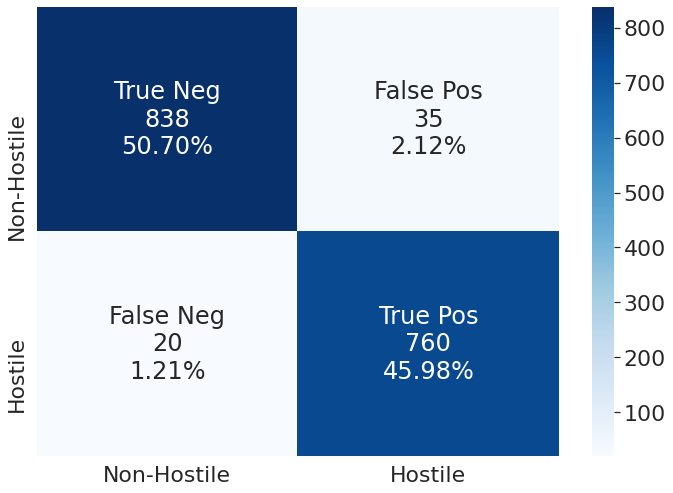}
  \label{fig:sub1}
\end{subfigure}%
\begin{subfigure}{.5\textwidth}
  \centering
  \includegraphics[width=.8\linewidth]{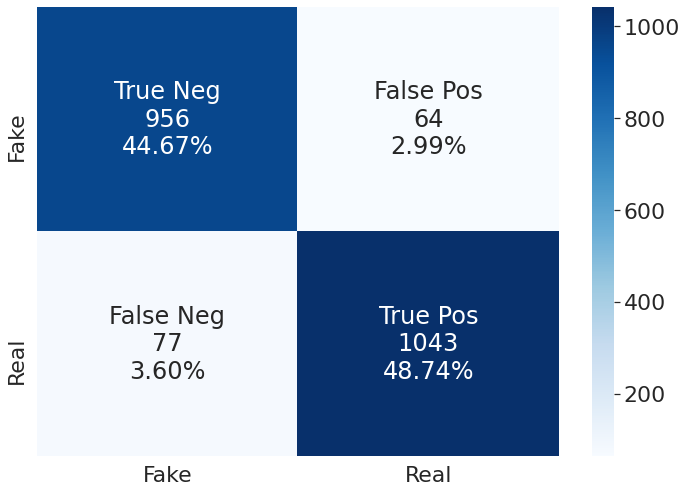}
  \label{fig:sub2}
  
\end{subfigure}
\caption{Confusion Matrix  of Hindi Hostility coarse-grain classification (left) \& English Fake News classification (right)}
\label{fig:confusion}
\end{figure}

\begin{table}[!htb]
    \begin{minipage}{.5\linewidth}
      \centering
        \begin{tabular}{|l|l|l|l|l|}
\hline
\textbf{Class Name} & \textbf{Precision}        & \textbf{Recall} & \textbf{F1} \\ \hline
Non-Hostile         & 0.98 & 0.96            & 0.97             \\ \hline
Hostile             & 0.96                      & 0.97            & 0.97                \\ \hline
W.avg/Total         & 0.97                      & 0.97            & 0.97                \\ \hline
\end{tabular}
    \end{minipage}%
    \begin{minipage}{.4\linewidth}
      \centering
        \begin{tabular}{|c|c|c|c|c|}
\hline
\textbf{Class name} & \textbf{precision} & \textbf{Recall} & \textbf{F1} & \textbf{Support}         \\ \hline
Defamation          & 0.30               & 0.72            & 0.42        & 169 \\ \hline
Fake                & 0.67               & 0.84            & 0.77        & 334 \\ \hline
Hate                & 0.44               & 0.80            & 0.57        & 237 \\ \hline
Offensive           & 0.46               & 0.79            & 0.59        & 219 \\ \hline
W.avg/Total         & 0.50               & 0.80            & 0.62        & 959                      \\ \hline
\end{tabular}
    \end{minipage}
    
    \caption{The Coarse-grain performance (left) \& the Fine-grain performance of the best Ensemble Model (right).}
    \label{hindi-expet-table}
\end{table}



\subsection{Hindi Hostility Detection Results}

\textbf{Baseline Classifiers}: We see that the FastText based feature combined with LSTM yields the best result, equally matched by the mBERT based embeddings. TF-IDF based models perform poorly, due to the lack of contextual understanding of the text. (Table \ref{tab:hate-results})\\
\textbf{Ablation Study}: In Table \ref{tab:hate-results}, the addition of metadata (m1,m2,m3) increases the F1 score by 1\% in coarse-grain and 2\% in fine-grain respectively. But when m1 and m3 are given together F1 increases by 4\%, suggesting a better combination. When all are provided together to the model, an increase of 7\% F1 can be observed in test data yielding to best F1 in fine-grain. Based on Table \ref{tab:hate-results} we observe that the fake tweet are easier to detect as compared to other classes. We also see that, due to the abusive language detector(m1) as a feature, we are able to detect the offensive tweets better.\\
\textbf{Proposed Model}: As an individual model, Hindi BERT Embedding with BERT model yields the best result for coarse-grain classification and FastText-LSTM with metadata yields best result on fine-grain classification on test data. Our best score is achieved on the Hindi-Hostility task, where we secured 5th rank out of 45 teams using an ensemble model. From Figure \ref{fig:confusion} we can see our Ensemble model predicts only 20 out of 1653 tweets as False Negative and 35 out of 1653 detected as a false negative, yielding 0.97 F1.

%% file: Conclusion.tex
\section{Conclusion}\label{Conclusion}

In this paper, we address hostility and fake news detection on the \textit{Devanagari} (Hindi language) script using an ensemble model, coupled with abusive language detector and metadata to attain an F1 score of 0.97. Furthermore, we also attained an F1 score of 0.93 on the English fake news detection task using a model based on Word2Vec embeddings, metadata, and entity features. Our results highlight that representation learning models augmented with data specific rules and features outperform the vanilla deep learning models.

Both the tasks attempted in our paper are far from solved and bringing in modalities like, images or videos, social network structures, user-based embeddings, event-based embeddings and more could help improve the current model. Furthermore, as the manual curation of such datasets is laborious and prone to errors, methods from semi-supervised learning and weak supervision could be leveraged.